\documentclass{article}
\usepackage[accepted]{icml2020}                  
\usepackage{hyperref}       
\usepackage{microtype}      
\usepackage{graphicx}
\usepackage{booktabs}       
\usepackage{url}            
\usepackage{amsfonts}       
\usepackage{nicefrac}       
\usepackage{lipsum}
\usepackage{amsmath}
\usepackage{subcaption}     
\usepackage{enumitem}
\usepackage{wrapfig}
\usepackage{amssymb}
\usepackage{epstopdf}
\usepackage{bbold}

\usepackage[usenames,dvipsnames]{xcolor}
\definecolor{myblue2}{HTML}{4682B4}
\definecolor{myred}{HTML}{FF5733}

\icmltitlerunning{Negative sampling in semi-supervised learning}

\begin{document}

\twocolumn[

\icmltitle{Negative sampling in semi-supervised learning}

\begin{icmlauthorlist}
\icmlauthor{John Chen}{rice}
\icmlauthor{Vatsal Shah}{ut}
\icmlauthor{Anastasios Kyrillidis}{rice}
\end{icmlauthorlist}

\icmlcorrespondingauthor{John Chen}{johnchen@rice.edu}
\icmlaffiliation{rice}{Department of Computer Science, Rice University, Houston, Texas USA}
\icmlaffiliation{ut}{Department of Electrical and Computer Engineering, University of Texas at Austin, Austin, Texas USA}

\icmlkeywords{Semi-supervised Learning, Negative Sampling}

\vskip 0.3in
]\printAffiliationsAndNotice{} 
\begin{abstract}
    We introduce Negative Sampling in Semi-Supervised Learning ($\text{NS}^3\text{L}$), a simple, fast, easy to tune algorithm for semi-supervised learning (SSL). 
    $\text{NS}^3\text{L}$ is motivated by the success of negative sampling/contrastive estimation. We demonstrate that adding the $\text{NS}^3\text{L}$ loss to state-of-the-art SSL algorithms, such as the Virtual Adversarial Training (VAT), significantly improves upon vanilla VAT and its variant, VAT with Entropy Minimization. 
    By adding the $\text{NS}^3\text{L}$ loss to MixMatch, the current state-of-the-art approach on semi-supervised tasks, we observe significant improvements over vanilla MixMatch.
    We conduct extensive experiments on the CIFAR10, CIFAR100, SVHN and STL10 benchmark datasets. Finally, we perform an ablation study for $\text{NS}^3\text{L}$ regarding its hyperparameter tuning.
    
\end{abstract}

\section{Introduction}
Deep learning has been hugely successful in areas such as image classification \citep{krizhevsky2012imagenet, he2016deep, zagoruyko2016wide, huang2017densely} and speech recognition \citep{sak2014long,sercu2016very}, where a large amount of labeled data is available. 
However, in practice it is often prohibitively expensive to create a large, high quality labeled dataset, due to lack of time, resources, or other factors. 
For example, the ImageNet dataset---which consists of 3.2 million labeled images in 5247 categories---took nearly two and half years to complete with the aid of Amazon's Mechanical Turk \citep{deng2009imagenet}.
Some medical tasks may require months of preparation, expensive hardware, the collaboration of many experts, and often are limited by the number of participants \citep{miotto2016deep}. 
As a result, it is desirable to exploit unlabeled data to aid the training of deep learning models.

This form of learning is semi-supervised learning \citep{chapelle2006semi} (SSL). 
Unlike supervised learning, the aim of SSL is to leverage unlabeled data, in conjunction with labeled data, to improve performance. 
SSL is typically evaluated on labeled datasets where a certain proportion of labels have been discarded.
There have been a number of instances in which SSL is reported to achieve performance close to purely supervised learning \citep{laine2017temporal, miyato2017virtual, tarvainen2017mean, berthelot2019mixmatch}, where the purely supervised learning model is trained on the much larger whole dataset. 
However, despite significant progress in this field, it is still difficult to quantify when unlabeled data may aid the performance except in a handful of cases \citep{balcan2005pac,ben2008does,kaariainen2005generalization,niyogi2013manifold,rigollet2007generalization,singh2009unlabeled, wasserman2008statistical}.

\textit{In this work, we restrict our attention to SSL algorithms which add a loss term to the neural network loss.}
These algorithms are the most flexible and practical given the difficulties in hyperparameter tuning in the entire model training process, in addition to achieving the state-of-the-art performance. 

We introduce \textit{Negative Sampling in Semi-Supervised Learning ($\text{NS}^3\text{L}$)}: a simple, fast, easy to tune SSL algorithm, motivated by negative sampling/contrastive estimation \citep{mikolov2013dist, smith2005contrastive}. 
In negative sampling/contrastive estimation, in order to train a model on unlabeled data, we exploit \textit{implicit} negative evidence, originating from the unlabeled samples: 
Using negative sampling, we seek for good models that discriminate a supervised example from its
neighborhood, comprised of unsupervised examples, assigned with a random (and potentially wrong) class. 
Stated differently, the learner learns that not only the supervised example is good, but that the same example is locally optimal in the space of examples, and that alternative examples are inferior. 
With negative sampling/contrastive estimation, instead of explaining and exploiting all of the data (that is not available during training), the model implicitly must only explain why the observed, supervised example is better than its unsupervised neighbors.

Overall, $\text{NS}^3\text{L}$ adds a loss term to the learning objective, and is shown to improve performance simply by doing so to other state-of-the-art SSL objectives. 
Since modern datasets often have a large number of classes \citep{imagenet}, we are motivated by the observation that \textit{it is often much easier to label a sample with a class or classes it is not, as opposed to the one class it is,} exploiting ideas from negative sampling/contrastive estimation \citep{mikolov2013dist, smith2005contrastive}.

\paragraph{Key Contributions.} Our findings can be summarized as follows:
\begin{itemize}[leftmargin=*]
  \item[$i)$] We propose a new SSL algorithm, which is easy to tune, and improves SSL performance of other state of the art algorithms across a wide range of reasonable hyperparameters, simply by adding the $\text{NS}^3\text{L}$ loss in their objective. 
  \item[$ii)$] Adding the $\text{NS}^3\text{L}$ loss to a variety of losses, including Virtual Adversarial Training (VAT) \citep{miyato2017virtual}, $\Pi$ model, and MixMatch \citep{berthelot2019mixmatch}, we observe improved performance compared to vanilla alternatives as well as the addition of Pseudo-Labeling or Entropy Minimization, for the standard SSL benchmarks of SVHN, CIFAR10, and CIFAR100. 
  \item[$iii)$] Adding the $\text{NS}^3\text{L}$ loss to the state-of-the-art SSL algorithm, i.e., the MixMatch procedure \citep{berthelot2019mixmatch}, $\text{NS}^3\text{L}$ combined with MixMatch produces superior performance for the standard SSL benchmarks of SVHN, CIFAR10 and STL-10.
\end{itemize}
Namely, adding the $\text{NS}^3\text{L}$ loss to existing SSL algorithms is an easy way to improve performance, and requires limited extra computational resources for hyperparameter tuning, since it is interpretable, fast, and sufficiently easy to tune.

\begin{figure*}
    \centering
      \begin{subfigure}[b]{0.60\linewidth}
        \includegraphics[width=\linewidth]{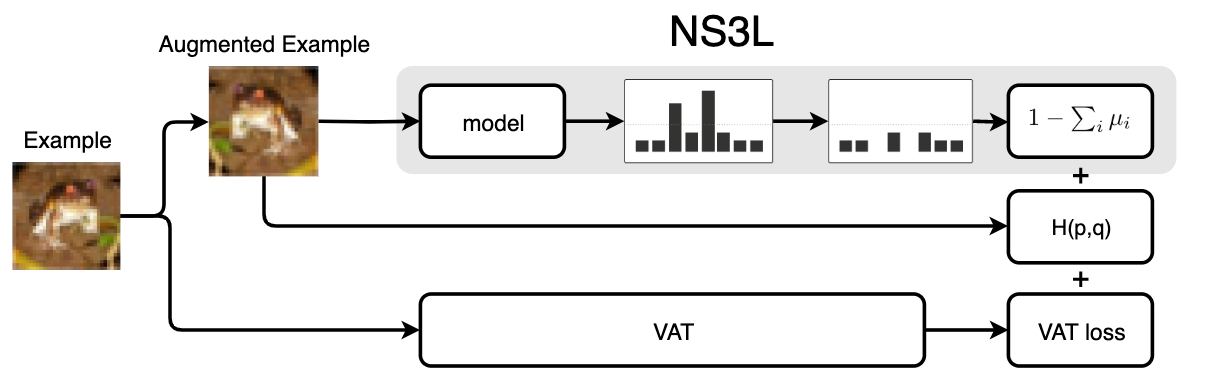}
        \label{fig:ptblstm25epochssgdcomp}
      \end{subfigure} \hspace{-0.2cm}
      \begin{subfigure}[b]{0.38\linewidth}
        \includegraphics[width=\linewidth]{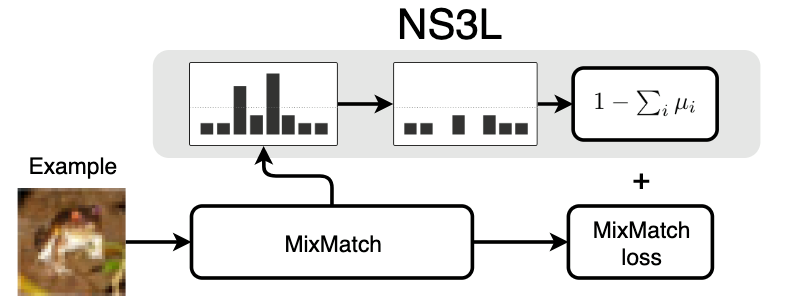}
        \label{fig:vaemnist200pochssgdcomp}
      \end{subfigure}
    \caption{Left: Diagram of NS$^3$L with VAT. For NS$^3$L, an augmented example is fed into the model, which outputs a probability for each class. A threshold $T$ is used to determine classes with sufficiently low probability, and these classes are fed into the NS$^3$L loss. The NS$^3$L loss is combined with the existing VAT loss and Cross Entropy loss. Right: Similar diagram of NS$^3$L with MixMatch; the NS$^3$L loss is combined with the existing MixMatch loss.}
    \label{fig:ns3l_diagram}
\end{figure*}

\section{Related Work}

In this paper, we restrict our attention to a subset of SSL algorithms which add a loss to the supervised loss function. 
\textit{These algorithms tend to be more practical in terms of hyperparameter tuning} \citep{berthelot2019mixmatch}. 
There are a number of SSL algorithms not discussed in this paper, following and as mentioned in \citep{berthelot2019mixmatch}: including "transductive" models \citep{joachim1999trans,joachim2003trans,gammerman1998learn}, graph-based methods \citep{zhu2003semi,bengio2006label}, and generative modeling \citep{joachim2003trans,belkin2002laplacian, salak2007deepbelief, coates2011encoding, goodfellow2011spike, kingma2014semi, odena2016semi, pu2016variational, salimans2016improved}. For a comprehensive overview of SSL methods, refer to \cite{chapelle2006semi}, or \cite{zhu2003semi}.

\subsection{Consistency Regularization}
Consistency regularization applies data augmentation to semi-supervised learning with the following intuition: 
Small perturbations for each sample should not significantly change the output of the network. 
This is usually achieved by minimizing some distance measure between the output of the network, with and without perturbations in the input. 
The most straightforward distance measure is the mean squared error used by the $\Pi$ model \citep{laine2017temporal, sajjadi2016regularization}. 
The $\Pi$ model adds the distance term $d(f_{\theta}(x),f_{\theta}(\hat{x}))$, where $\hat{x}$ is the result of a stochastic perturbation to $x$, to the supervised classification loss as a regularizer, with some weight.

Mean teacher \citep{tarvainen2017mean} observes the potentially unstable target prediction over the course of training with the $\Pi$ model approach, and proposes a prediction function, parameterized by an exponential moving average of model parameter values. 
Mean teacher adds $d(f_{\theta}(x),f_{\theta '}(x))$, where $\theta '$ is an exponential moving average of $\theta$, to the supervised classification loss with some weight. However, the stochastic perturbation used in these methods are domain specific. 

\subsection{Virtual Adversarial Training}
Virtual Adversarial Training \citep{miyato2017virtual} (VAT) approximates perturbations to be applied over the input to most significantly affect the output class distribution, inspired by adversarial examples \citep{goodfellow2015explain,szegedy2014intriguing}. 
VAT computes an approximation of the perturbation as: 
\begin{align*}
        r &\sim \mathcal{N}\left(0, \tfrac{\xi}{\sqrt{\mathrm{dim}(x)}} I\right)\\
        g &= \nabla_r d\left(f_{\theta}(x), f_{\theta}(x + r)\right)\\
        r_{\text{adv}} &= \epsilon \cdot g/\|g\|_2 
\end{align*}
where $x$ is an input data sample, $\text{dim}(\cdot)$ is its dimension, $d$ is a non-negative function that measures the
divergence between two distributions, $\xi$ and $\epsilon$ are scalar hyperparameters. 
Consistency regularization is then used to minimize the distance between the output of the network, with and without the perturbations in the input.
Since we follow the work in \cite{oliver2018realistic} almost exactly, we select the best performing consistency regularization SSL method in that work, VAT, for comparison and combination with $\text{NS}^3\text{L}$ for non-Mixup SSL; Mixup procedure will be described later.

\subsection{Entropy minimization}
The goal of entropy minimization \citep{grandvalet2005entmin} is to discourage the decision boundary from passing near samples where the network produces low-confidence predictions. 
One way to achieve this is by adding a simple loss term to minimize the entropy for unlabeled data $x$ with total $K$ classes: $-\sum_{k=1}^{K} \mu_{xk} \log \mu_{xk}.$
Entropy minimization on its own has not demonstrated competitive performance in SSL, however it can be combined with VAT for stronger results \citep{miyato2017virtual, oliver2018realistic}. We include entropy minimization with VAT in our experiments.

\subsection{Pseudo-Labeling}
Pseudo-Labeling \citep{lee2013pseudo} is a simple and easy to tune method which is widely used in practice. For a particular sample, it requires only the probability value of each class, the output of the network, and labels the sample with a class if the probability value crosses a certain threshold. The sample is then treated as a labeled sample with the standard supervised loss function. Pseudo-Labeling is closely related to entropy minimization, but only enforces low-entropy predictions for predictions which are already low-entropy. We emphasize here that the popularity of Pseudo-Labeling is likely due to its simplicity and limited extra cost for hyperparameter search.

\subsection{SSL with modern data augmentation techniques}
Mixup \citep{zhang2017mixup} combines pairs of samples and their one-hot labels $(x_1, y_1), (x_2, y_2)$ as in: $x' = \lambda x_1 + (1 - \lambda) x_2, ~~y' = \lambda y_1 + (1 - \lambda) y_2$,
where $\lambda \sim \texttt{Beta}(\alpha, \alpha)$,
to produce a new sample $(x', y')$ with $\alpha$ being a hyperparameter. Mixup is a form of regularization which encourages the neural network to behave linearly between training examples, justified by Occam's Razor \citep{zhang2017mixup}. 
In SSL, the labels $y_1, y_2$ are typically the predicted labels by a neural network with some processing steps. 

Applying Mixup to SSL led to Interpolation Consistency Training (ICT) \citep{verma2019ict} and MixMatch \citep{berthelot2019mixmatch}, which significantly improved upon previous results with SSL on the standard benchmarks of CIFAR10 and SVHN. 
ICT trains the model $f_{\theta}$ to output predictions similar to a mean-teacher $f_{\theta '}$, where $\theta '$ is an exponential moving average of $\theta$. 
Namely, on unlabeled data, ICT encourages $f_{\theta} (\texttt{Mixup}(x_i, x_j)) \approx \texttt{Mixup}(f_{\theta '}(x_i), f_{\theta '}(x_j))$. 

MixMatch applies a number of processing steps for labeled and unlabeled data on each iteration and mixes both labeled and unlabeled data together. The final loss is given by $\mathcal{L} = \mathcal{L}_{\text{supervised}} + \lambda_3 \mathcal{L}_{\text{ubsupervised}}$, where 
\begin{align*}
    \mathcal{X}', \mathcal{U}' & = \texttt{MixMatch}(\mathcal{X}, \mathcal{U}, E, A, \alpha)
    \end{align*}
     \begin{align*}
    \mathcal{L}_{\text{supervised}} & = \frac{1}{|\mathcal{X}'|} \sum_{i_1 \in \mathcal{X}'} \sum_{k=1}^{K} y_{i_1 k} \log \mu_{i_1 k}\\
    \mathcal{L}_{\text{unsupervised}} & = \frac{1}{K|\mathcal{U}'|} \sum_{i_2 \in \mathcal{U}'} \sum_{k=1}^{K} ( y_{i_2 k} - \mu_{i_2 k} ) ^2
\end{align*}
where $\mathcal{X}$ is the labeled data $\{x_{i_1}, y_{i_1}\}_{i_1 = 1}^n$, $\mathcal{U}$ is the unlabeled data $\{x_{i_2}^{u}\}_{i_2 = 1}^{n_u}$, $\mathcal{X}'$ and $\mathcal{U}'$ are the output samples labeled by MixMatch, and $E$, $A$, $\alpha$, $\lambda_3$ are hyperparameters. 
Given a batch of labeled and unlabeled samples, MixMatch applies $A$ data augmentations on each unlabeled sample $x_{i_2}$, averages the predictions across the $A$ augmentations,  
\begin{align*}
    p = \frac{1}{A} \sum_{a=1}^{A} f_{\theta}(\texttt{Augment}(x_{i_2}^{u}))
\end{align*}
and applies temperature sharpening, 
\begin{align*}
    \texttt{Sharpen}(p, E)_k := \frac{p_k^{1/E}}{\sum_{k=1}^{K} p_k^{1/E}},
\end{align*}
to the average prediction. $A$ is typically 2 in practice, and $E$ is 0.5. The unlabeled data is labeled with this sharpened average prediction. 

Let the collection of labeled unlabeled data be $\mathcal{\widehat{U}}$. Standard data augmentation is applied to the originally labeled data and let this be denoted $\mathcal{\widehat{X}}$.
Let $\mathcal{W}$ denote the shuffled collection of $\mathcal{\widehat{U}}$ and $\mathcal{\widehat{X}}$. MixMatch alters Mixup by adding a max operation: $\lambda \sim \texttt{Beta}(\alpha, \alpha), ~~\lambda ' = \max(\lambda, 1 - \lambda)$;
it then produces $\mathcal{X}' = \texttt{Mixup}(\mathcal{\widehat{X}}_{i_1} , W_{i_1})$ and $\mathcal{U}' = \texttt{Mixup}(\mathcal{\widehat{U}}_{i_2} , W_{i_2 + |\mathcal{\widehat{X}}|})$.

Since MixMatch performs the strongest empirically, we select MixMatch as the best performing Mixup-based SSL method for comparison and combination with $\text{NS}^3\text{L}$. We make a note here that more recently there is also work on applying stronger data augmentation \citep{xie2019unsupervised}.

\section{Negative Sampling in Semi-Supervised Learning}

In this section, we provide the pseudo-code for the Negative Sampling with Semi-Supervised Learning ($\text{NS}^3\text{L}$) algorithm in Algorithm \ref{alg:ns3l_loss}. $\text{NS}^3\text{L}$ assigns a random label to an unsupervised sample as long as the probability of that random label being correct is low. Adding $\text{NS}^3\text{L}$ to any existing algorithms allows us to achieve significant performance improvements. We first provide the mathematical motivation behind  $\text{NS}^3\text{L}$ followed by intuition of why  $\text{NS}^3\text{L}$ works using a simple toy example in $1D$.

\subsection{Mathematical Motivation}

Let the set of labeled samples be denoted as $\{ x_i, y_i\}_{i=1}^n$, $x_i$ being the input and $y_i$ being the associated label, and the set of unlabeled samples be denoted as $\{x_i^u\}_{i=1}^{n_u}$, each with unknown correct label $y_i^{u}$. 
For the rest of the text, we will consider the cross-entropy loss, which is one of the most widely used loss functions for classification.
The objective function for cross entropy loss over the labeled examples is:
\begin{align*}
\mathcal{L}\left(\{x_i, y_i\}_{i = 1}^n \right) = - \tfrac{1}{n} \sum_{i = 1}^n \sum_{k = 1}^K y_{ik} \log \mu_{ik}, 
\end{align*}
where there are $n$ labeled samples, $K$ classes, $y_{ik} = \mathbb{1}_{k = y_i}$ is the identity operator that equals 1 when $k = y_i$, and $\mu_{ik}$ is the output of the classifier for sample $i$ for class $k$. 

\begin{figure*}[!t]
    \centering
    \includegraphics[scale=0.2]{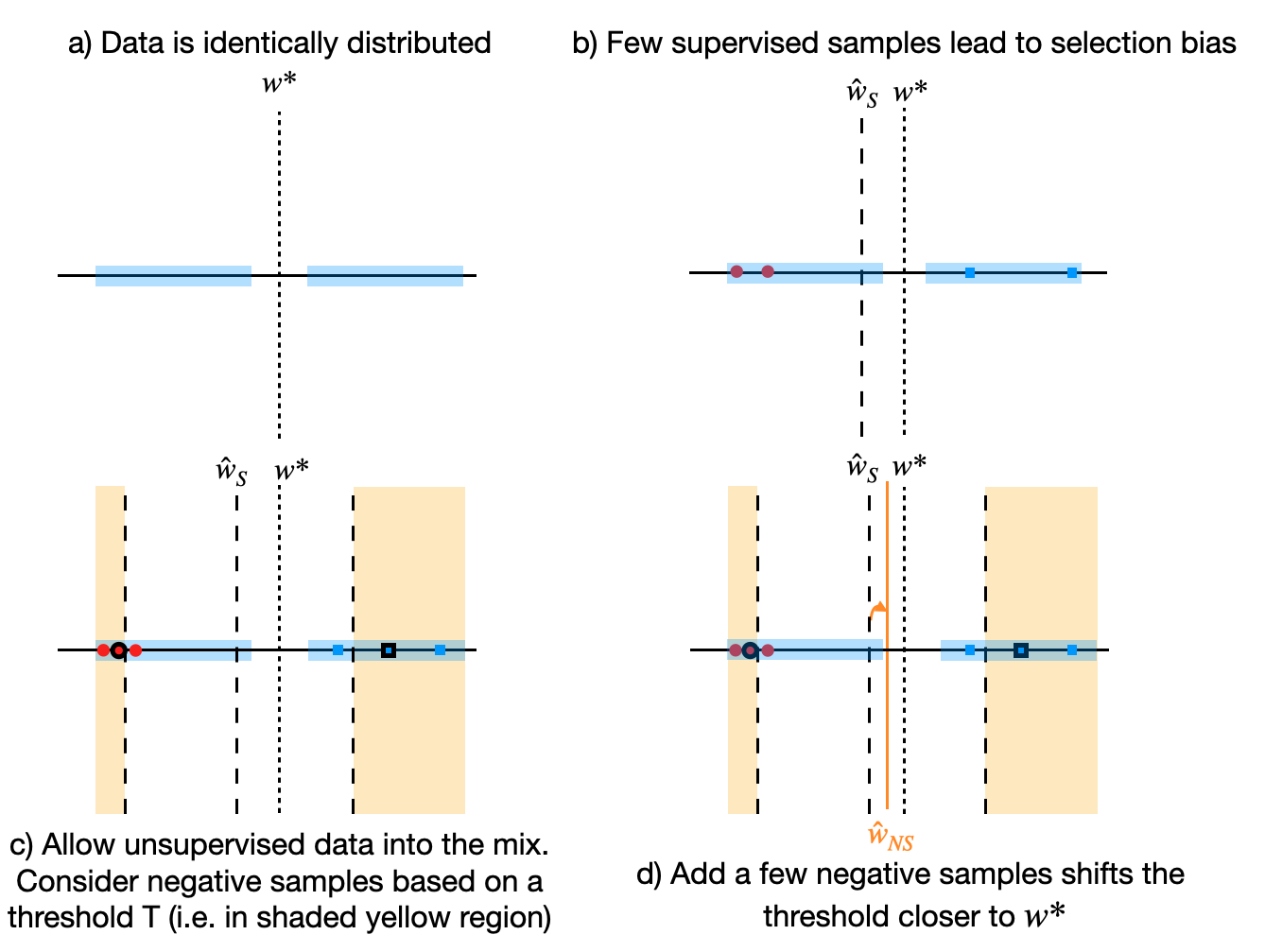}
    \caption{A toy example illustrating the effectiveness of Negative Sampling in Semi supervised learning}
    \label{fig:toy1}
\end{figure*}
For the sake of simplicity, we will perform the following relabeling: for all $i \in [n_u]$,  $x_{i+n}=x_i^u$ and $y_{i+n}=y_i^u$. 
In the hypothetical scenario where the labels for the unlabeled data are known and for $w$ the parameters of the model, 
the likelihood would be:
\begin{align*}
\mathbb{P}\left[ \{y_i\}_{i = 1}^{n + n_u}~|~\{x_i\}_{i = 1}^{n + n_u}, w \right] \\
&\hspace{-2cm}= \prod_{i = 1}^{n + n_u} \mathbb{P}\left[ y_i~|~x_i, w \right] = \prod_{i = 1}^{n + n_u} \prod_{k = 1}^K \mu_{ik}^{y_{ik}} \: , \\
&\hspace{-2cm}= \left( \prod_{i_1 = 1}^{n} \prod_{k = 1}^K \mu_{i_1k}^{y_{i_1k}}\right) \cdot \left (\prod_{i_2 = 1}^{n_u} \prod_{k = 1}^K \mu_{i_2k}^{y_{i_2k}^u} \right) 
\end{align*}
Observe that, $\prod_{k = 1}^K \mu_{i_2k}^{y_{i_2k}^u} = 1 - \sum_{j: y_{i_2 j} \neq 1} \mu_{i_2 j}$,
which follows from the definition of the quantities $\mu_{:}$ that represent a probability distribution and, consequently, sum up to one.

Taking negative logarithms allows us to split the loss function into two components: $i)$ the supervised part and $ii)$ the unsupervised part. The log-likelihood loss function can now be written as follows:
\begin{align*}
\mathcal{L}\left( \{ x_i, y_i\}_{i = 1}^{n + n_u} \right) &= -  \underbrace{\tfrac{1}{n} \sum_{i_1 = 1}^n \sum_{k = 1}^K y_{ik} \log \mu_{ik}}_{:= \text{supervised part}} \\
&- \underbrace{\tfrac{1}{n_u} \sum_{i_2 = 1}^{n_u} \log \left(1 - \sum_{j \neq \text{True label}} \mu_{i_2j} \right)}_{:= \text{unsupervised part}} 
\end{align*}

While the true labels need to be known for the unsupervised part to be accurate, we draw ideas from negative sampling/contrastive estimation \citep{mikolov2013dist, smith2005contrastive}:
i.e., for each unlabeled example in the unsupervised part, we randomly assign $P$ labels from the set of labels; see also Appendix A.
These $P$ labels indicate classes that \textit{the sample does not belong to}: as the number of labels in the task increase, the probability of including the correct label in the set of $P$ labels is small.
The way labels are selected could be uniformly at random or by using Nearest Neighbor search, or even based on the output probabilities of the network, where with high probability the correct label is not picked. 

\begin{algorithm}[!t]
    \centering
    \caption{\textsc{$\text{NS}^3\text{L}$}}\label{alg:ns3l_loss}
    \begin{algorithmic}[1]
        \STATE \textbf{Input}: Mini batch size $B$, batch of examples $x_b$ and their predicted vector of label probabilities $\hat{y}_b$ using the output of the classifier $\{x_b, \hat{y}_b\}_{b = 1}^{B}$, threshold $T$.
        \STATE $\mathcal{L}_{\texttt{$\text{NS}^3\text{L}$}} = 0$.
        \FOR{$b = 1, \dots, B$}
        \STATE $\mathbb{1}_{\hat{y}_b'} = \texttt{isTrue}(\hat{y}_b < T)$.
        \STATE $\mathcal{L}_{\texttt{$\text{NS}^3\text{L}$}} = \mathcal{L}_{\texttt{$\text{NS}^3\text{L}$}} - \log\left(1 - \sum_{k=1}^{K} \mathbb{1}_{\hat{y}_{b k}'} \mu_{b k}\right)$.
        \ENDFOR  \\
        \STATE \textbf{Return} $\frac{1}{B} \mathcal{L}_{\texttt{$\text{NS}^3\text{L}$}}$
    \end{algorithmic}
\end{algorithm}
The approach above assumes the use of the full dataset, both for the supervised and unsupervised parts. 
In practice, more often than not we train models based on stochastic gradient descent, and we implement a mini-batch variant of this approach with different batch sizes $B_1$ and $B_2$ for labeled and unlabeled data, respectively. 
Particularly, for the supervised mini-batch of size $B_1$ for labeled data, the objective term is approximated as: 
\begin{align*}
\tfrac{1}{n} \sum_{i_1 = 1}^n \sum_{k = 1}^K y_{ik} \log \mu_{ik} \approx \tfrac{1}{|B_1|} \sum_{i_1 \in B_1} \sum_{k = 1}^K y_{ik} \log \mu_{ik}.
\end{align*}
The unsupervised part with mini-batch size of $B_2$ and $\text{NS}^3\text{L}$ loss, where each unlabeled sample is connected with $P_{i_2}$ hopefully incorrect labels, is approximated as:
\begin{align*}
\tfrac{1}{n_u} \sum_{i_2 = 1}^{n_u} \log \big(1 - \sum_{j \neq \text{True label}} \mu_{i_2j} \big) \\
&\hspace{-2cm}\approx \tfrac{1}{|B_2|} \sum_{i_2 \in B_2} \log \left(1 - \sum_{j = 1}^{P_{i_2}} \mu_{i_2 j} \right)
\end{align*}

Based on the above, our $\text{NS}^3\text{L}$ loss looks as follows: 
\begin{align*}
\hat{\mathcal{L}}_{B_1, B_2}\left( \{x_i, y_i\}_{i = 1}^{n + n_u} \right)&=
 -\tfrac{1}{|B_1|} \sum_{i_1 \in B_1} \sum_{k = 1}^K y_{ik} \log \mu_{ik} \\
 &- \underbrace{\tfrac{1}{|B_2|} \sum_{i_2 \in B_2} \log \left(1 - \sum_{j = 1}^{P_{i_2}} \mu_{i_2 j} \right)}_{:= \text{$\text{NS}^3\text{L}$ loss}}
\end{align*}
Thus, the $\text{NS}^3\text{L}$ loss is just an additive loss term that can be easily included in many existing SSL algorithms, as we show next. 
For clarity, a pseudocode implementation of the algorithm where negative labels are identified by the label probability being below a threshold $T$, as the output of the classifier or otherwise, is given in Algorithm \ref{alg:ns3l_loss}. 



\subsection{Intuition}

Our aim is to illustrate how our simple idea aids the task of learning with unlabeled data. 
We will consider a simple example in 1D (Figure \ref{fig:toy1}), where we assume binary classification with cross-entropy loss for simplicity.

Let $w^\star$ denote the separating hyperplane and assume that the data lies uniformly on either side of $w^\star$, indicated by the shaded blue region (Figure \ref{fig:toy1}a). 
Without loss of generality, let the points on the left and right of the hyperplane have the labels $1$ and $0$, respectively. 
Our aim is to recover $w^\star$. 

It is possible for the labeled examples to have a selection bias \citep{chawla2005learning} (for example certain images of cats are easier to label than others); assume that this property leads the algorithm to converge to $\hat{w}$; Figure \ref{fig:toy1}b. 
However, in the SSL setting, we do have access to a large number of unlabeled examples. 
How can we utilize it to improve our prediction?

Consider one of the highlighted samples $(x_u)$ (red dot with black boundary in Figure \ref{fig:toy1}c). 
Let us assume its underlying true label is $1$. 
The key difference in both approaches is that in inductive SSL \citep{chapelle2006semi,zhu2003semi} we make a gradient update by labeling any point in the shaded yellow region as the predicted label while in negative sampling we make a gradient update by labeling the same point as not 0. 
Both these algorithms only perform updates only if we are certain about the label. 

Now, let us compare the gradients of a sample using the classical inductive SSL approach and negative sampling.
 
\begin{align*}
\text{Inductive SSL}: &\nabla \mathcal{L}\left(\{x_u\}\right) = -(1-\mu_{u}) x_u \\
\text{NS$^3$L}: &\nabla \mathcal{L}\left(\{x_u\} \right) =  \mu_{u} x_u
\end{align*}
From the equations above, it is clear that $\text{NS}^3\text{L}$ and Inductive SSL push the gradients in opposite directions. 
The gradient updates of supervised samples align with the gradient updates of the unsupervised samples labeled using Inductive SSL.
However, that is not the case for $\text{NS}^3\text{L}$. 
Since the unsupervised data samples come from a uniform distribution, it is more likely that we will pick more ``negative'' samples from the class on the right (intersection of yellow and blue shaded regions). These negative samples have a bias to the right side of the plane ultimately bringing back the separating hyper-plane closer to $w^\star$ (Figure \ref{fig:toy1}d).

\section{Experiments} \label{experiments_section}
We use the codebase from \cite{berthelot2019mixmatch} for experiments involving MixMatch, and otherwise use the codebase from \cite{oliver2018realistic}. We make the distinction due to the existence of some experimental differences, and this is the best way to reproduce the reported performances. Namely, \cite{berthelot2019mixmatch} differs from \cite{oliver2018realistic} in that it evaluates an exponential moving average of the model parameters, as opposed to using a learning rate decay schedule, and uses weight decay.


\subsection{Experimental Setup} \label{experimentSetup}

Following \cite{oliver2018realistic}, the model employed is the standard Wide ResNet (WRN) \citep{zagoruyko2016wide} with depth 28 and width 2, batch normalization \citep{ioffe2015batch}, and leaky ReLU activations \citep{maas2013rectifier}. 
The optimizer is the Adam optimizer \citep{kingma2014adam}. 
The batch size is 100, half of which are labeled and half are unlabeled. 
Standard procedures for regularization, data augmentation, and preprocessing are followed. 

We use the standard training data/validation data split for SVHN, with 65,932 training images and 7,325 validation images. 
All but 1,000 examples are turned ''unlabeled''. 
Similarly, we use the standard training/data validation data split for CIFAR10, with 45,000 training images and 5,000 validation images. 
All but 4,000 labels are turned ''unlabeled''. 
We also use the standard data split for CIFAR100, with 45,000 training images and 5,000 validation images. 
All but 10,000 labels are turned ''unlabeled''. 

Hyperparameters are optimized to minimize validation error; test error is reported at the point of lowest validation error. 
We select hyperparameters which perform well for both SVHN and CIFAR10. 
After selecting hyperparameters on CIFAR10 and SVHN, we run the same hyperparameters with practically no further tuning on CIFAR100 to determine the ability of each method to generalize to new datasets. 
Since VAT and VAT + EntMin use different hyperparameters for CIFAR10 and SVHN, we use those tuned for CIFAR10 for the CIFAR100 dataset. 
For $\text{NS}^3\text{L}$, $\text{NS}^3\text{L}$ + $\Pi$ model, $\text{NS}^3\text{L}$ + VAT, we divide the threshold $T$ by 10 since there are 10x classes in CIFAR100. 
We run 5 seeds for all cases.

Since models are typically trained on CIFAR10 \citep{krizhevsky2009cifar10} and SVHN \citep{netzer2011svhn} for fewer than the 500,000 iterations (1,000 epochs) \citep{oliver2018realistic}, we make the only changes of reducing the total iterations to 200,000, warmup period \citep{tarvainen2017mean} to 50,000, and iteration of learning rate decay to 130,000. 
All other methodology follows that work \citep{oliver2018realistic}. 

For MixMatch experiments, we follow the methodology of \cite{berthelot2019mixmatch} and continue to use the same model described above. 
Since the performance of MixMatch is particularly strong using only a small number of labeled samples, we also include experiments for SVHN with all but 250 labels discarded, and CIFAR10 with all but 250 labels discarded, in addition to the previously mentioned experiments. 
We also include experiments on STL10, a dataset designed for SSL, which has 5,000 labeled images and 100,000 unlabeled images drawn from a slightly different distribution than the labeled data. 
All but 1,000 labels are discarded for STL10. The median of the last 20 checkpoints' test error is reported, following \cite{berthelot2019mixmatch}. Note that we reduce the training epochs of STL10 significantly in interest of training time. 
All other methodology follows the work of MixMatch.

\subsection{Baseline Methods}
For baseline methods, we consider Pseudo-Labeling, due to its simplicity on the level of $\text{NS}^3\text{L}$, and MixMatch and VAT for its performance, in addition to VAT + Entropy Minimization and VAT + Psuedo-Labeling. 
We also include $\Pi$ model and omit Mean Teacher, although we follow the experiments of \cite{oliver2018realistic} and both produce worse performance than VAT. 
The supervised baseline is trained on the remaining labeled data after some labels have been removed. 
We generally follow the tuned hyperparameters in the literature and do not observe noticeable gains from further hyperparameter tuning.

\begin{table*}[htbp]
\centering \setlength\tabcolsep{4.9pt}
\begin{footnotesize}
\caption{Test errors achieved by various SSL approaches on the standard benchmarks of CIFAR10, with all but 4,000 labels removed, SVHN, with all but 1,000 labels removed, and CIFAR100, with all but 10,000 labels removed. "Supervised" refers to using only 4,000, 1,000, and 10,000 labeled samples from CIFAR10, SVHN, and CIFAR100 respectively without any unlabeled data. VAT refers to Virtual Adversarial Training.} 
\begin{tabular}{c|cccccccc} \toprule
    Dataset & Supervised & PL & $\text{NS}^3\text{L}$ & VAT & VAT + EntMin & $\Pi$ model & $\Pi$ + $\text{NS}^3\text{L}$ &  VAT + $\text{NS}^3\text{L}$\\ \midrule
    CIFAR10 & 20.76 $\pm$ .28 & 17.56 $\pm$ .29 & 16.03 $\pm$ .05 & 14.72 $\pm$ .23 & 14.34 $\pm$ .18 &  17.12 $\pm$ .19 & 16.06 $\pm$ .21 &  \textcolor{myred}{\textbf{13.94}} $\pm$ .10 \\
    SVHN & 12.39 $\pm$ .53 & 7.70 $\pm$ .22 & 6.52 $\pm$ .22 & 6.20 $\pm$ .11 & 6.10 $\pm$ .02 & 8.48 $\pm$ .15 &  7.98 $\pm$ .18  &  \textcolor{myred}{\textbf{5.51}} $\pm$ .14 \\
    CIFAR100 & 48.26 $\pm$ .25 & 46.91 $\pm$ .31 & 46.34 $\pm$ .37 & 44.38 $\pm$ .56 & 43.92 $\pm$ .44 & 47.87 $\pm$ .34 &  46.98 $\pm$ .41 &  \textcolor{myred}{\textbf{43.70}} $\pm$ .19 \\
 \bottomrule
\end{tabular}
\label{tableResults}
\end{footnotesize}
\end{table*}
\subsection{Implementation of $\text{NS}^3\text{L}$} \label{implementation1}
We implement $\text{NS}^3\text{L}$ using the output probabilities of the network with the unlabeled samples, namely 
\begin{align*}
    \mathcal{L}_{\texttt{$\text{NS}^3\text{L}$}} = \texttt{$\text{NS}^3\text{L}$}(\{x_{i_2}, \mu_{i_2}\}_{i_2=1}^{B} , T).
\end{align*}
The performance of $\text{NS}^3\text{L}$ with random negative sampling assignment or Nearest Neighbor-based assignment is given in Section \ref{sec:alternative} in the appendix.
We label a sample with negative labels for the classes whose probability value falls below a certain threshold. 
We then simply add the $\text{NS}^3\text{L}$ loss to the existing SSL loss function. 
Using $\text{NS}^3\text{L}$ on its own gives
\begin{align*}
    \mathcal{L} = \mathcal{L}_{\text{supervised}} + \lambda_1\mathcal{L}_{\texttt{$\text{NS}^3\text{L}$}}
\end{align*}
for some weighting $\lambda_1$. For adding $\text{NS}^3\text{L}$ to VAT, this gives
\begin{align*}
    \mathcal{L} = \mathcal{L}_{\text{supervised}} + \lambda_2 \mathcal{L}_{\text{VAT}} + \lambda_1 \mathcal{L}_{\texttt{$\text{NS}^3\text{L}$}}
\end{align*}
for some weighting $\lambda_i, i \in \{1, 2\}$. This is applied similarly to the $\Pi$ model. The weighting is a common practice in SSL, also used in MixMatch and VAT + Entropy Minimization. This is the simplest form of $\text{NS}^3\text{L}$ and we believe there are large gains to be made with more complex methods of choosing the negative labels.

Recall that MixMatch outputs $\mathcal{X}', \mathcal{U}' = \texttt{MixMatch}(\mathcal{X}, \mathcal{U}, T, A, \alpha)$ collections of samples with their generated labels. 
We label each sample $x_i \in \mathcal{X}' \bigcup \mathcal{U}'$ with negative labels for the classes whose generated probability value falls below a certain threshold. 
We then simply add the $\text{NS}^3\text{L}$ loss to the existing SSL loss function, computing the $\text{NS}^3\text{L}$ loss using the probability outputs of the network as usual. Namely,
\begin{align*}
    \mathcal{X}', \mathcal{U}' & = \texttt{MixMatch}(\mathcal{X}, \mathcal{U}, E, A, \alpha) \\
    \mathcal{L}_{\text{supervised}} & = \frac{1}{|\mathcal{X}'|} \sum_{i_1 \in \mathcal{X}'} \sum_{k=1}^{K} y_{i_1 k} \log \mu_{i_1 k} \\
    \mathcal{L}_{\text{unsupervised}} & = \frac{1}{K|\mathcal{U}'|} \sum_{i_2 \in \mathcal{U}'} \sum_{k=1}^{K} ( y_{i_2 k} - \mu_{i_2 k} ) ^2
\end{align*}
\begin{align*}
    \mathcal{L}_{\texttt{$\text{NS}^3\text{L}$}} & = \texttt{$\text{NS}^3\text{L}$}(\mathcal{X}' \bigcup \mathcal{U}' , T) \\
    \mathcal{L} & = \mathcal{L}_{\text{supervised}} + \lambda_3 \mathcal{L}_{\text{unsupervised}} + \lambda_1 \mathcal{L}_{\texttt{$\text{NS}^3\text{L}$}}
\end{align*}

\subsection{Results}

\begin{table*}[!h]
\centering
\caption{Test errors achieved by various SSL approaches on top of VAT on the standard benchmarks of CIFAR10, with all but 4,000 labels removed, and CIFAR100, with all but 10,000 labels removed. VAT, EntMin and PL refer to Virtual Adversarial Training, Entropy Minimization, and Pseudo-Labeling respectively.}
\begin{tabular}{c|cccccccc} \toprule
    Dataset & VAT & VAT + EntMin & VAT + PL & VAT + $\text{NS}^3\text{L}$\\ \midrule
    CIFAR10 & 14.72 $\pm$ .23 & 14.34 $\pm$ .18 & 14.15 $\pm$ .14  & \textcolor{myred}{\textbf{13.94}} $\pm$ .10 \\
    CIFAR100 & 44.38 $\pm$ .56 & 43.92 $\pm$ .44 &   43.93 $\pm$ .33   & \textcolor{myred}{\textbf{43.70}} $\pm$ .19 \\
 \bottomrule
\end{tabular}
\label{tableResultsVAT}
\end{table*}

We follow the practice in \cite{oliver2018realistic} and use the same hyperparameters for plain $\text{NS}^3\text{L}$ and $\text{NS}^3\text{L}$ as added to other losses, e.g. $\text{NS}^3\text{L}$ + VAT, for both CIFAR10 and SVHN. After selecting hyperparameters on CIFAR10 and SVHN, we run almost the exact same hyperparameters with little further tuning on CIFAR100, where the threshold $T$ is divided by 10 since there are 10x classes in CIFAR100.

For MixMatch experiments, we follow the practice of \cite{berthelot2019mixmatch} and tune $\text{NS}^3\text{L}$ separately for each dataset. 
MixMatch + $\text{NS}^3\text{L}$ only takes marginally longer runtime than MixMatch on its own. 
The learning rate is fixed.
\paragraph{CIFAR10:} We evaluate the accuracy of each method with 4,000 labeled samples and 41,000 unlabeled samples, as is standard practice. The results are given in Table \ref{tableResults}. Further results comparing the addition of Entropy Minimization, Pseudo-Labeling and $\text{NS}^3\text{L}$ are given in Table \ref{tableResultsVAT}. MixMatch results are given in Table \ref{tableMixmatchCifar10Results}.
For $\text{NS}^3\text{L}$, we use a threshold $T = 0.04$, learning rate of 6e-4, and $\lambda_1 = 1$. 
Identical hyperparameters are used for $\Pi$ model + $\text{NS}^3\text{L}$. For VAT + $\text{NS}^3\text{L}$, we use a shared learning rate of 6e-4 and reduce $\lambda_1$ from 1 to 0.3, which is identical to $\lambda_2$. We perform extensive hyperparameter tuning for VAT + PL. For MixMatch, as in \cite{berthelot2019mixmatch}, we use $\alpha = 0.75$ and $\lambda_3 = 75$. 
For $\text{NS}^3\text{L}$ + MixMatch, we use a threshold of $T = 0.05$ and a coefficient of $\lambda_1 = 5$ for 250 labeled samples and $\lambda_1 = 10$ for 4,000 labeled samples.
All other settings remain as is optimized individually. 

We created 5 splits of the number of labeled samples, each with a different seed. Each model is trained on a different split and test error is reported with mean and standard deviation. 
We find that $\text{NS}^3\text{L}$ performs reasonably well and significantly better than Pseudo-Labeling, over a 1.5\% improvement. 
A significant gain over all algorithms is attained by adding the $\text{NS}^3\text{L}$ loss to the VAT loss. 
VAT + $\text{NS}^3\text{L}$ achieves almost a 1\% improvement over VAT, and is about 0.5\% better than VAT + EntMin and VAT + PL. We also find that adding $\text{NS}^3\text{L}$ immediately improves the performance of MixMatch, with a 2\% improvement with 250 labeled samples and a small improvement for 4,000 samples. 
The 250 labeled samples case may be the more interesting case since it highlights the sample efficiency of the method. 
This underscores the flexibility of $\text{NS}^3\text{L}$ to improve existing methods. 

\begin{table}[!h]
\centering
\caption{Test errors achieved by MixMatch and MixMatch + $\text{NS}^3\text{L}$ on the standard benchmark of CIFAR10, with all but 250 labels removed and all but 4,000 labels removed.}
\begin{tabular}{c|cc} \toprule
    CIFAR10 & 250 & 4,000 \\ \midrule
    MixMatch & 14.49 $\pm$ 1.60 & 7.05 $\pm$ 0.10 \\
    Mixmatch + $\text{NS}^3\text{L}$ & \textcolor{myred}{\textbf{12.48}} $\pm$ 1.21 & \textcolor{myred}{\textbf{6.92}} $\pm$ 0.12\\
 \bottomrule
\end{tabular}
\label{tableMixmatchCifar10Results}
\end{table}

\paragraph{SVHN:} We evaluate the accuracy of each method with 1,000 labeled samples and 64,932 unlabeled samples, as is standard practice. 
The results are shown in Table \ref{tableResults}. MixMatch results are shown in Table \ref{tableMixmatchSVHNResults}.
We use the same hyperparameters for $\text{NS}^3\text{L}$, $\Pi$ model + $\text{NS}^3\text{L}$ and VAT + $\text{NS}^3\text{L}$ as in CIFAR10. For MixMatch ollowing the literature, we use $\alpha = 0.75$ and $\lambda_3 = 250$. 
For $\text{NS}^3\text{L}$ + MixMatch, we again use a threshold of $T = 0.05$ and a coefficient of $\lambda_1 = 2$ for both 250 labeled samples and 1,000 labeled samples. 

Again, 5 splits are created, each with a different seed. 
Each model is trained on a different split and test error is reported with mean and standard deviation. 
Here, $\text{NS}^3\text{L}$ achieves competitive learning rate with VAT, 6.52\% versus 6.20\%, and is significantly better than Pseudo-Labeling, at 7.70\%. By combining $\text{NS}^3\text{L}$ with VAT, test error is further reduced by a notable margin, almost 1\% better than VAT alone and more than 0.5\% better than VAT + EntMin. 

By adding $\text{NS}^3\text{L}$ to MixMatch, the model achieves almost the same test error with 250 labeled samples than it does using only MixMatch on 1,000 labeled samples. 
In other words, in this case applying $\text{NS}^3\text{L}$ improves performance almost equivalent to having 4x the amount of labeled data. 
In the cases of 250 labeled samples and 1,000 labeled samples, adding $\text{NS}^3\text{L}$ to MixMatch improves performance by 0.4\% and 0.15\% respectively, achieving state-of-the-art results.

\begin{table}[!h]
\centering
\caption{Test errors achieved by MixMatch and MixMatch + $\text{NS}^3\text{L}$ on the standard benchmark of SVHN, with all but 250 labels removed and all but 1,000 labels removed.}
\begin{tabular}{c|cc} \toprule
    SVHN & 250 & 1,000 \\ \midrule
    MixMatch & 3.75 $\pm$ 0.09 & 3.28 $\pm$ 0.11 \\
    Mixmatch + $\text{NS}^3\text{L}$ & \textcolor{myred}{\textbf{3.38}} $\pm$ 0.08 & \textcolor{myred}{\textbf{3.14}} $\pm$ 0.11\\
 \bottomrule
\end{tabular}
\label{tableMixmatchSVHNResults}
\end{table}

\paragraph{STL10:} We evaluate the accuracy of MixMatch and MixMatch + $\text{NS}^3\text{L}$ with 1,000 labeled samples and 100,000 unlabeled samples. The results are given in Table \ref{tableMixmatchSTL10Results}. 
Following the literature, we use $\alpha=0.75$ and $\lambda_3=50$. For $\text{NS}^3\text{L}$, we again use a threshold of $T = 0.05$ and $\lambda_1 = 2$. We trained the model for a significantly fewer epochs than in \cite{berthelot2019mixmatch}, however even in this case $\text{NS}^3\text{L}$ can improve upon MixMatch, reducing test error slightly.

\begin{table}[!h]
\centering
\caption{Test errors achieved by MixMatch and MixMatch + $\text{NS}^3\text{L}$ on the standard benchmark of STL10, with all but 1,000 labels removed.}
\begin{tabular}{c|cc} \toprule
    STL10 & 1,000 \\ \midrule
    MixMatch & 22.20 $\pm$ 0.89 \\
    Mixmatch + $\text{NS}^3\text{L}$ & \textcolor{myred}{\textbf{21.74}} $\pm$ 0.33 \\
 \bottomrule
\end{tabular} 
\label{tableMixmatchSTL10Results}
\end{table}

\paragraph{CIFAR100;} We evaluate the accuracy of each method with 10,000 labeled samples and 35,000 unlabeled samples, as is standard practice. 
The results are given in Table \ref{tableResults}. 
For $\text{NS}^3\text{L}$, we use a threshold $T = 0.04 / 10 = 0.004$, learning rate of 6e-4, and $\lambda_1 = 1$, following the settings in CIFAR10 and SVHN. 
For VAT + $\text{NS}^3\text{L}$ in CIFAR100, we use a shared learning rate of 3e-3 and $\lambda_1 = 0.3$, $\lambda_2 = 0.6$. 

As before, we created 5 splits of 10,000 labeled samples, each with a different seed, and each model is trained on a different split. 
Test error is reported with mean and standard deviation. 
$\text{NS}^3\text{L}$ is observed to improve 0.6\% test error over Pseudo-Labeling and adding $\text{NS}^3\text{L}$ to VAT reduces test error slightly and achieves the best performance. 
This suggests that EntMin and $\text{NS}^3\text{L}$ boosts VAT even with little hyperparameter tuning, and perhaps should be used as default.
We note that the performance of SSL methods can be sensitive to hyperparameter tuning, and minor hyperparameter tuning may improve performance greatly. 
\emph{Due to VAT performing additional forward and backwards passes, $\text{NS}^3\text{L}$ alone runs more than 2x faster than VAT.}

\begin{figure*}[!t]
  \centering
  \begin{subfigure}[b]{0.33\linewidth}
    \includegraphics[width=\linewidth]{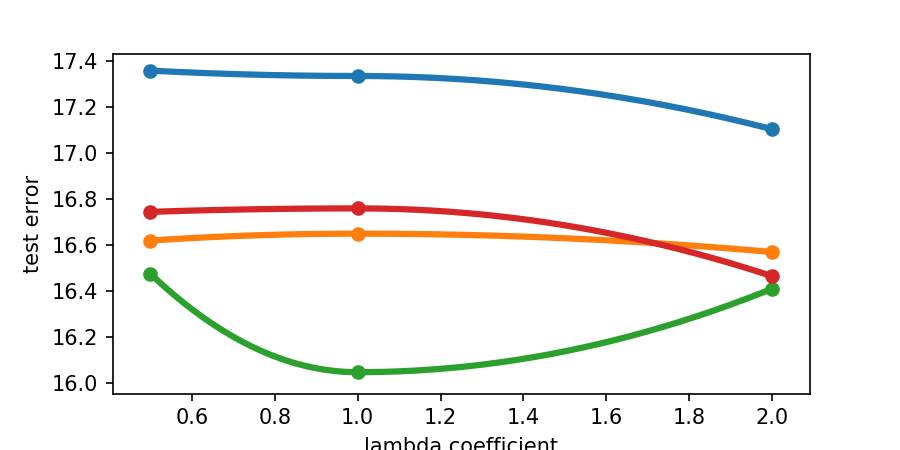}
    \label{fig:parameterSensitivityNS3L}
  \end{subfigure}
  \begin{subfigure}[b]{0.33\linewidth}
    \includegraphics[width=\linewidth]{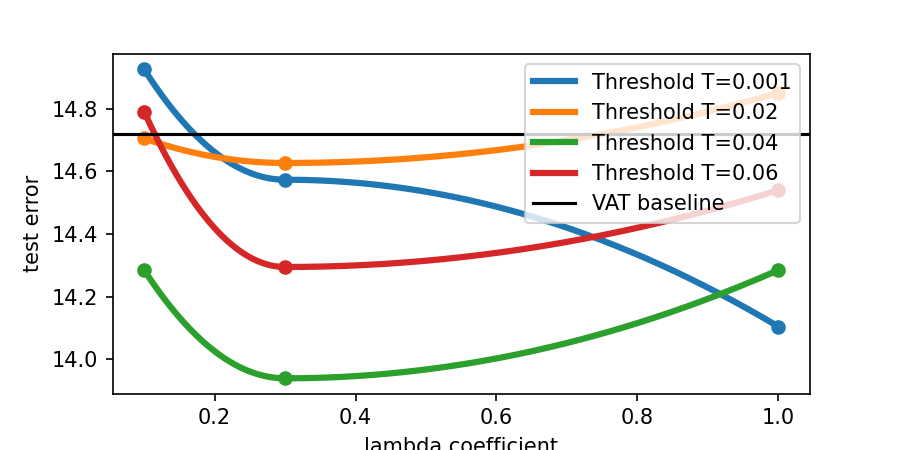}
    \label{fig:parameterSensitivityNS3LVAT}
  \end{subfigure}
  \begin{subfigure}[b]{0.33\linewidth}
    \includegraphics[width=\linewidth]{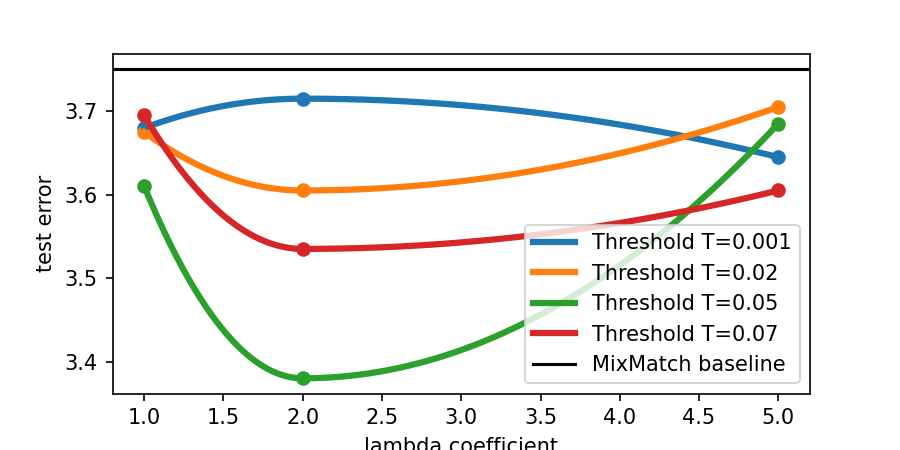}
    \label{fig:parameterSensitivityNS3LMixmatch}
  \end{subfigure}
  \caption{Parameter sensitivity study. Left: Test errors achieved by $\text{NS}^3\text{L}$ on the standard benchmark of CIFAR10, with all but 4,000 labels removed. Middle: Test errors achieved by VAT + $\text{NS}^3\text{L}$ on the standard benchmark of CIFAR10, with all but 4,000 labels removed. Right row: Test errors achieved by Mixmatch + $\text{NS}^3\text{L}$ on the standard benchmark of SVHN, with all but 250 labels removed.}
  \label{fig:sensitivity}
\end{figure*}
\section{Parameter Sensitivity}
We provide experimental results on the sensitivity of $\text{NS}^3\text{L}$ with respect to the threshold parameter $T$ and the weighting parameter $\lambda_1$. We use the CIFAR10 dataset with all but 4,000 labels removed for $\text{NS}^3\text{L}$ and VAT + $\text{NS}^3\text{L}$. We use the SVHN dataset with all but 250 labels removed for MixMatch + $\text{NS}^3\text{L}$. We fix all other optimal parameters given in Section \ref{experiments_section}. Results are given in Figure \ref{fig:sensitivity}, where 4 values of threshold $T$ and 3 values of weighting parameter $\lambda_1$ are selected. We interpolated the result for better readability.

Referring to Figure \ref{fig:sensitivity}, the optimal $\lambda_1$ depends on the setting and is affected when used simultaneously with VAT or Mixmatch. E.g., the optimal $\lambda_1$ for $\text{NS}^3\text{L}$ on CIFAR10 with all but 4,000 labels removed varies from approximately 1, when added to the existing cross entropy loss alone, to 0.3, when added to the cross entropy loss and VAT with a coefficient of 0.3. When added to Mixmatch on SVHN with all but 250 labels removed, the optimal $\lambda_1$ is closer to 2.

The performance is more sensitive to the threshold $T$, and an optimal threshold $T \approx 0.04$ appears to hold empirically across settings, and we note that the datasets are all of 10 classes.
Referring to Table \ref{tableResults} and Table \ref{tableMixmatchSVHNResults}, we see a clear improvement by adding NS3L, even when it is poorly tuned.

\section{Conclusion}
With simplicity, speed, and ease of tuning in mind, we proposed \emph{Negative Sampling in Semi-Supervised Learning} ($\text{NS}^3\text{L}$), a semi-supervised learning method inspired by negative sampling, which simply adds a loss function. 
We demonstrate the effectiveness of $\text{NS}^3\text{L}$ when combined with existing SSL algorithms, producing the overall best result for non-Mixup-based SSL, by combining $\text{NS}^3\text{L}$ with VAT, and Mixup-based SSL, by combining $\text{NS}^3\text{L}$ with MixMatch. 
We show improvements across a variety of tasks with only a minor increase in training time.

\bibliographystyle{icml2020}
\bibliography{references}

\begin{thebibliography}{51}
\providecommand{\natexlab}[1]{#1}
\providecommand{\url}[1]{\texttt{#1}}
\expandafter\ifx\csname urlstyle\endcsname\relax
  \providecommand{\doi}[1]{doi: #1}\else
  \providecommand{\doi}{doi: \begingroup \urlstyle{rm}\Url}\fi

\bibitem[Balcan \& Blum(2005)Balcan and Blum]{balcan2005pac}
Balcan, M.-F. and Blum, A.
\newblock A pac-style model for learning from labeled and unlabeled data.
\newblock In \emph{International Conference on Computational Learning Theory},
  pp.\  111--126. Springer, 2005.

\bibitem[Belkin \& Niyogi(2002)Belkin and Niyogi]{belkin2002laplacian}
Belkin, M. and Niyogi, P.
\newblock Laplacian eigenmaps and spectral techniques for embedding and
  clustering.
\newblock In \emph{Advances in Neural Information Processing Systems}, 2002.

\bibitem[Ben-David et~al.(2008)Ben-David, Lu, and P{\'a}l]{ben2008does}
Ben-David, S., Lu, T., and P{\'a}l, D.
\newblock Does unlabeled data provably help? worst-case analysis of the sample
  complexity of semi-supervised learning.
\newblock In \emph{COLT}, pp.\  33--44, 2008.

\bibitem[Bengio et~al.(2006)Bengio, Delalleau, and Le~Roux]{bengio2006label}
Bengio, Y., Delalleau, O., and Le~Roux, N.
\newblock Label propagation and quadratic criterion.
\newblock \emph{MIT Press}, 2006.

\bibitem[Berthelot et~al.(2019)Berthelot, Carlini, Goodfellow, Papernot, and
  Raffel]{berthelot2019mixmatch}
Berthelot, D., Carlini, N., Goodfellow, I., Papernot, Nicolas~Oliver, A., and
  Raffel, C.
\newblock Mixmatch: A holistic approach to semi-supervised learning.
\newblock \emph{arXiv preprint arXiv:1905.02249}, 2019.

\bibitem[Chapelle \& Scholkopf(2006)Chapelle and Scholkopf]{chapelle2006semi}
Chapelle, O. and Scholkopf, B.
\newblock Semi-supervised learning.
\newblock \emph{MIT Press}, 2006.

\bibitem[Chawla \& Karakoulas(2005)Chawla and Karakoulas]{chawla2005learning}
Chawla, N.~V. and Karakoulas, G.
\newblock Learning from labeled and unlabeled data: An empirical study across
  techniques and domains.
\newblock \emph{Journal of Artificial Intelligence Research}, 23:\penalty0
  331--366, 2005.

\bibitem[Coates \& Ng(2011)Coates and Ng]{coates2011encoding}
Coates, A. and Ng, A.~Y.
\newblock The importance of encoding versus training with sparse coding and
  vector quantization.
\newblock In \emph{International Conference on Machine Learning}, 2011.

\bibitem[Deng et~al.(2009)Deng, Dong, Socher, Li, Li, and
  Fei-Fei]{deng2009imagenet}
Deng, J., Dong, W., Socher, R., Li, L.-J., Li, K., and Fei-Fei, L.
\newblock Imagenet: A large-scale hierarchical image database.
\newblock In \emph{2009 IEEE conference on computer vision and pattern
  recognition}, pp.\  248--255. Ieee, 2009.

\bibitem[Gammerman et~al.(1998)Gammerman, Vovk, and Vapnik]{gammerman1998learn}
Gammerman, A., Vovk, V., and Vapnik, V.
\newblock Learning by transduction.
\newblock In \emph{Proceedings of the Fourteenth Conference on Uncertainty in
  Artificial Intelligence}, 1998.

\bibitem[Goodfellow et~al.(2011)Goodfellow, Courville, and
  Bengio]{goodfellow2011spike}
Goodfellow, I.~J., Courville, A., and Bengio, Y.
\newblock Spike-and-slab sparse coding for unsupervised feature discovery.
\newblock \emph{NIPS Workshop on Challenges in Learning Hierarchical Models},
  2011.

\bibitem[Goodfellow et~al.(2015)Goodfellow, Shlens, and
  Szegedy]{goodfellow2015explain}
Goodfellow, I.~J., Shlens, J., and Szegedy, C.
\newblock Explaining and harnessing adversarial examples.
\newblock In \emph{International Conference on Learning Representations}, 2015.

\bibitem[Grandvalet \& Bengio(2005)Grandvalet and Bengio]{grandvalet2005entmin}
Grandvalet, Y. and Bengio, Y.
\newblock Semi-supervised learning by entropy minimization.
\newblock In \emph{Advances in Neural Information Processing Systems}, 2005.

\bibitem[He et~al.(2016)He, Zhang, Ren, and Sun]{he2016deep}
He, K., Zhang, X., Ren, S., and Sun, J.
\newblock Deep residual learning for image recognition.
\newblock In \emph{Proceedings of the IEEE conference on computer vision and
  pattern recognition}, pp.\  770--778, 2016.

\bibitem[Huang et~al.(2017)Huang, Liu, Van Der~Maaten, and
  Weinberger]{huang2017densely}
Huang, G., Liu, Z., Van Der~Maaten, L., and Weinberger, K.~Q.
\newblock Densely connected convolutional networks.
\newblock In \emph{Proceedings of the IEEE conference on computer vision and
  pattern recognition}, pp.\  4700--4708, 2017.

\bibitem[Ioffe \& Szegedy(2015)Ioffe and Szegedy]{ioffe2015batch}
Ioffe, S. and Szegedy, C.
\newblock Batch normalization: Accelerating deep network training.
\newblock In \emph{International Conference on Machine Learning}, 2015.

\bibitem[Joachims(1999)]{joachim1999trans}
Joachims, T.
\newblock Transductive inference for text classification using support vector
  machines.
\newblock In \emph{International Conference on Machine Learning}, 1999.

\bibitem[Joachims(2003)]{joachim2003trans}
Joachims, T.
\newblock Transductive learning via spectral graph partitioning.
\newblock In \emph{International Conference on Machine Learning}, 2003.

\bibitem[K{\"a}{\"a}ri{\"a}inen(2005)]{kaariainen2005generalization}
K{\"a}{\"a}ri{\"a}inen, M.
\newblock Generalization error bounds using unlabeled data.
\newblock In \emph{International Conference on Computational Learning Theory},
  pp.\  127--142. Springer, 2005.

\bibitem[Kingma \& Ba(2014)Kingma and Ba]{kingma2014adam}
Kingma, D.~P. and Ba, J.
\newblock Adam: A method for stochastic optimization.
\newblock \emph{arXiv preprint arXiv:1412.6980}, 2014.

\bibitem[Kingma et~al.(2014)Kingma, Mohamed, Rezende, and
  Welling]{kingma2014semi}
Kingma, D.~P., Mohamed, S., Rezende, D.~J., and Welling, M.
\newblock Semisupervised learning with deep generative models.
\newblock In \emph{Advances in Neural Information Processing Systems}, 2014.

\bibitem[Krizhevsky(2009)]{krizhevsky2009cifar10}
Krizhevsky, A.
\newblock Learning multiple layers of features from tiny images.
\newblock \emph{Technical report}, 2009.

\bibitem[Krizhevsky et~al.(2012)Krizhevsky, Sutskever, and
  Hinton]{krizhevsky2012imagenet}
Krizhevsky, A., Sutskever, I., and Hinton, G.~E.
\newblock Imagenet classification with deep convolutional neural networks.
\newblock In \emph{Advances in neural information processing systems}, pp.\
  1097--1105, 2012.

\bibitem[Laine \& Aila(2017)Laine and Aila]{laine2017temporal}
Laine, S. and Aila, T.
\newblock Temporal ensembling for semi-supervised learning.
\newblock In \emph{International Conference on Learning Representations}, 2017.

\bibitem[Lee(2013)]{lee2013pseudo}
Lee, D.-H.
\newblock Pseudo-label: The simple and efficient semi-supervised learning
  method for deep neural networks.
\newblock \emph{ICML Workshop on Challenges in Representation Learning}, 2013.

\bibitem[Maas \& Ng(2013)Maas and Ng]{maas2013rectifier}
Maas, Andrew~L., H. A.~Y. and Ng, A.~Y.
\newblock Rectifier nonlinearities improve neural network acoustic models.
\newblock In \emph{International Conference on Machine Learning}, 2013.

\bibitem[Mikolov et~al.(2013)Mikolov, Sutskever, Chen, Corrado, and
  Dean]{mikolov2013dist}
Mikolov, T., Sutskever, I., Chen, K., Corrado, G., and Dean, J.
\newblock Distributed representations of words and phrases and their
  compositionality.
\newblock In \emph{Advances in Neural Information Processing Systems}, 2013.

\bibitem[Miotto et~al.(2016)Miotto, Li, Kidd, and Dudley]{miotto2016deep}
Miotto, R., Li, L., Kidd, B.~A., and Dudley, J.~T.
\newblock Deep patient: an unsupervised representation to predict the future of
  patients from the electronic health records.
\newblock \emph{Scientific reports}, 6:\penalty0 26094, 2016.

\bibitem[Miyato et~al.(2017)Miyato, Maeda, Koyama, and
  Ishii]{miyato2017virtual}
Miyato, T., Maeda, S.-i., Koyama, M., and Ishii, S.
\newblock Virtual adversarial training: a regularization method for supervised
  and semi-supervised learning.
\newblock \emph{arXiv preprint arXiv:1704.03976}, 2017.

\bibitem[Netzer et~al.(2011)Netzer, Wang, Coates, Bissacco, Wu, and
  Ng]{netzer2011svhn}
Netzer, Y., Wang, T., Coates, A., Bissacco, A., Wu, B., and Ng, A.~Y.
\newblock Reading digits in natural images with unsupervised feature learning.
\newblock \emph{NIPS Workshop on Deep Learning and Unsupervised Feature
  Learning}, 2011.

\bibitem[Niyogi(2013)]{niyogi2013manifold}
Niyogi, P.
\newblock Manifold regularization and semi-supervised learning: Some
  theoretical analyses.
\newblock \emph{The Journal of Machine Learning Research}, 14\penalty0
  (1):\penalty0 1229--1250, 2013.

\bibitem[Odena(2016)]{odena2016semi}
Odena, A.
\newblock Semi-supervised learning with generative adversarial networks.
\newblock \emph{arXiv preprint arXiv:1606.01583}, 2016.

\bibitem[Oliver et~al.(2018)Oliver, Odena, Raffel, Cubuk, and
  Goodfellow]{oliver2018realistic}
Oliver, A., Odena, A., Raffel, C., Cubuk, E.~D., and Goodfellow, I.~J.
\newblock Realistic evaluation of deep semi-supervised learning algorithms.
\newblock \emph{arXiv preprint arXiv:1804.09170}, 2018.

\bibitem[Pu et~al.(2016)Pu, Zhe, Henao, Yuan, Li, Stevens, and
  Carin]{pu2016variational}
Pu, Y., Zhe, G., Henao, R., Yuan, X., Li, C., Stevens, A., and Carin, L.
\newblock Variational autoencoder for deep learning of images, labels and
  captions.
\newblock In \emph{Advances in Neural Information Processing Systems}, 2016.

\bibitem[Rigollet(2007)]{rigollet2007generalization}
Rigollet, P.
\newblock Generalization error bounds in semi-supervised classification under
  the cluster assumption.
\newblock \emph{Journal of Machine Learning Research}, 8\penalty0
  (Jul):\penalty0 1369--1392, 2007.

\bibitem[Russakovsky et~al.(2014)Russakovsky, Deng, Su, Krause, Satheesh, Ma,
  Huang, Karpathy, Khosla, Bernstein, Berg, and Li]{imagenet}
Russakovsky, O., Deng, J., Su, H., Krause, J., Satheesh, S., Ma, S., Huang, Z.,
  Karpathy, A., Khosla, A., Bernstein, M., Berg, A., and Li, F.-F.
\newblock Imagenet large scale visual recognition challenge.
\newblock \emph{arXiv preprint arXiv:1409.0575}, 2014.

\bibitem[Sajjadi et~al.(2016)Sajjadi, Javanmardi, and
  Tasdizen]{sajjadi2016regularization}
Sajjadi, M., Javanmardi, M., and Tasdizen, T.
\newblock Regularization with stochastic transformations and perturbations for
  deep semi-supervised learning.
\newblock In \emph{Advances in Neural Information Processing Systems}, 2016.

\bibitem[Sak et~al.(2014)Sak, Senior, and Beaufays]{sak2014long}
Sak, H., Senior, A., and Beaufays, F.
\newblock Long short-term memory recurrent neural network architectures for
  large scale acoustic modeling.
\newblock In \emph{Fifteenth annual conference of the international speech
  communication association}, 2014.

\bibitem[Salakhutdinov \& Hinton(2007)Salakhutdinov and
  Hinton]{salak2007deepbelief}
Salakhutdinov, R. and Hinton, Geoffrey, H.~E.
\newblock Using deep belief nets to learn covariance kernels for gaussian
  processes.
\newblock In \emph{Advances in Neural Information Processing Systems}, 2007.

\bibitem[Salimans et~al.(2016)Salimans, Goodfellow, Zaremba, Cheung, Radford,
  and Chen]{salimans2016improved}
Salimans, T., Goodfellow, I., Zaremba, W., Cheung, V., Radford, A., and Chen,
  X.
\newblock Improved techniques for training gans.
\newblock In \emph{Advances in Neural Information Processing Systems}, 2016.

\bibitem[Sercu et~al.(2016)Sercu, Puhrsch, Kingsbury, and LeCun]{sercu2016very}
Sercu, T., Puhrsch, C., Kingsbury, B., and LeCun, Y.
\newblock Very deep multilingual convolutional neural networks for {LVCSR}.
\newblock In \emph{2016 IEEE International Conference on Acoustics, Speech and
  Signal Processing (ICASSP)}, pp.\  4955--4959. IEEE, 2016.

\bibitem[Singh et~al.(2009)Singh, Nowak, and Zhu]{singh2009unlabeled}
Singh, A., Nowak, R., and Zhu, J.
\newblock Unlabeled data: Now it helps, now it doesn't.
\newblock In \emph{Advances in neural information processing systems}, pp.\
  1513--1520, 2009.

\bibitem[Smith \& Eisner(2005)Smith and Eisner]{smith2005contrastive}
Smith, N.~A. and Eisner, J.
\newblock Contrastive estimation: Training log-linear models on unlabeled data.
\newblock In \emph{Proceedings of the 43rd Annual Meeting on Association for
  Computational Linguistics}, pp.\  354--362. Association for Computational
  Linguistics, 2005.

\bibitem[Szegedy et~al.(2014)Szegedy, Zaremba, Sutskever, Bruna, Erhan,
  Goodfellow, and Fergus]{szegedy2014intriguing}
Szegedy, C., Zaremba, W., Sutskever, I., Bruna, J., Erhan, D., Goodfellow, I.,
  and Fergus, R.
\newblock Intriguing properties of neural networks.
\newblock In \emph{International Conference on Learning Representations}, 2014.

\bibitem[Tarvainen \& Valpola(2017)Tarvainen and Valpola]{tarvainen2017mean}
Tarvainen, A. and Valpola, H.
\newblock Mean teachers are better role models: Weight-averaged consistency
  targets improve semi-supervised deep learning results.
\newblock In \emph{Advances in Neural Information Processing Systems}, 2017.

\bibitem[Verma et~al.(2019)Verma, Lamb, Kannala, Bengio, and
  Lopez-Pas]{verma2019ict}
Verma, V., Lamb, A., Kannala, J., Bengio, Y., and Lopez-Pas, D.
\newblock Interpolation consistency training for semi-supervised learning.
\newblock \emph{arXiv preprint arXiv:1903.03825}, 2019.

\bibitem[Wasserman \& Lafferty(2008)Wasserman and
  Lafferty]{wasserman2008statistical}
Wasserman, L. and Lafferty, J.~D.
\newblock Statistical analysis of semi-supervised regression.
\newblock In \emph{Advances in Neural Information Processing Systems}, pp.\
  801--808, 2008.

\bibitem[Xie et~al.(2019)Xie, Dai, Hovy, Luong, and Le]{xie2019unsupervised}
Xie, Q., Dai, Z., Hovy, E., Luong, M.-T., and Le, Q.
\newblock Unsupervised data augmentation for consistency training.
\newblock \emph{arXiv preprint arXiv:1904.12848}, 2019.

\bibitem[Zagoruyko \& Komodakis(2016)Zagoruyko and
  Komodakis]{zagoruyko2016wide}
Zagoruyko, S. and Komodakis, N.
\newblock Wide residual networks.
\newblock \emph{arXiv preprint arXiv:1605.07146}, 2016.

\bibitem[Zhang et~al.(2017)Zhang, Cisse, Dauphin, and
  Lopez-Pas]{zhang2017mixup}
Zhang, H., Cisse, M., Dauphin, Y.~N., and Lopez-Pas, D.
\newblock mixup: Beyond empirical risk minimization.
\newblock \emph{arXiv preprint arXiv:1710.09412}, 2017.

\bibitem[Zhu et~al.(2003)Zhu, Ghahramani, and Lafferty]{zhu2003semi}
Zhu, X., Ghahramani, Z., and Lafferty, J.~D.
\newblock Semi-supervised learning using gaussian fields and harmonic
  functions.
\newblock In \emph{International Conference on Machine Learning}, 2003.

\end{thebibliography}

\newpage

\appendix

\newpage
\begin{appendix}

\section{Negative Sampling and its connection to \texttt{word2vec}} \label{negativeSamplingAppendix}
We  present the case of \texttt{word2vec} for negative sampling where the number of words and contexts is such that picking a random pair of (word, context) is with high probability not related.
To make the resemblance, let us describe the intuition behind \texttt{word2vec}. 
Here, the task is to relate words --represented as $w$-- with contexts --represented as $c$.
We can theoretically conceptualize words $w$ being related with $x$, and contexts being related to labels $y$.
The negative sampling by Mikolov et al., considers the following objective function: consider a pair $(w, c)$ of a word and a context. 
If this pair comes from valid data that correctly connects these two, then we can say that the data pair $(w, c)$ came from the true data distribution;
if this pair does otherwise, then we claim that $(w, c)$ does not come from the true distribution.

In math, we will denote by $\mathbb{P} \left[D = 1~|~w, c \right]$ as the probability that $(w, c)$ satisfies the first case, and $\mathbb{P}\left[D = 0~|~w, c\right]$ otherwise.
The paper models these probabilities as:
\begin{align*}
\mathbb{P}\left[D = 1 ~|~ w, c\right] = \frac{1}{1 + e^{-v_c^\top v_w}},
\end{align*}
where $v_c, v_w$ correspond to the vector representation of the context and word, respectively.

Now, in order to find good vector representations $\theta := \{v_c, v_w\}$ (we naively group all variables into $\theta$), given the data, we perform maximum log-likelihood as follows:
\begin{small}
\begin{align*}
\theta &= \arg\max_{\theta} \left( \prod_{(w, c) \in \mathcal{D}} \mathbb{P}\left[D = 1 ~|~ w, c, \theta \right] \right) \\ &\quad \quad \quad \quad \cdot \left( \prod_{(w, c) \notin \mathcal{D}} \mathbb{P}\left[D = 0 ~|~ w, c, \theta \right] \right) \\
          &= \arg\max_{\theta} \left( \prod_{(w, c) \in \mathcal{D}} \mathbb{P}\left[D = 1 ~|~ w, c, \theta \right] \right) \\ &\quad \quad \quad \quad  \cdot \left( \prod_{(w, c) \notin \mathcal{D}} \left(1 - \mathbb{P}\left[D = 1 ~|~ w, c, \theta \right]\right) \right) \\
          &= \arg\max_{\theta} \left( \sum_{(w, c) \in \mathcal{D}} \log \left(\mathbb{P}\left[D = 1 ~|~ w, c, \theta \right] \right) \right) \\ &\quad \quad \quad \quad  + \left( \sum_{(w, c) \notin \mathcal{D}} \log \left(1 - \mathbb{P}\left[D = 1 ~|~ w, c, \theta \right]\right) \right) \\
          &= \arg\max_{\theta} \left( \sum_{(w, c) \in \mathcal{D}} \log \left(\frac{1}{1 + e^{-v_c^\top v_w}} \right) \right) \\ &\quad \quad \quad \quad + \left( \sum_{(w, c) \notin \mathcal{D}} \log \left(\frac{1}{1 + e^{v_c^\top v_w}}\right) \right)
\end{align*}
\end{small}
Of course, we never take the whole dataset (whole corpus $\mathcal{D}$) and do gradient descent; rather we perform SGD by considering only a subset of the data for the first term:
\begin{align*}
\sum_{(w, c) \in \mathcal{D}} \log \left(\frac{1}{1 + e^{-v_c^\top v_w}} \right) \approx \sum_{(w, c) \in \text{mini-batch}} \log \left(\frac{1}{1 + e^{-v_c^\top v_w}} \right);
\end{align*}
Also, we cannot consider *every* data point not in the dataset; rather, we perform \emph{negative sampling} by selecting random pairs (according to some probability - this is important)---say $P$ pairs:
\begin{align*}
\left( \sum_{(w, c) \notin \mathcal{D}} \log \left(\frac{1}{1 + e^{v_c^\top v_w}}\right) \right) \approx \sum_{p = 1}^P \log \left(\frac{1}{1 + e^{-\tilde{v}_c^\top \tilde{v}_w}} \right),
\end{align*}
where the tildes represent the ``non-valid'' data.

\section{Alternative $\text{NS}^3\text{L}$ methods}{\label{sec:alternative}}
With computational efficiency in mind, we compare several methods of implementing $\text{NS}^3\text{L}$ in Table \ref{tableNS3LResults} on the F-MNIST dataset with a small Convolutional Neural Network. 
We split the F-MNIST dataset into a 2,000/58,000 labeled/unlabeled split and report validation error at the end of training. 
Specifically, we compare:

\begin{itemize} \itemsep-0.4em
  \item \texttt{Supervised:} trained only on the 2,000 labeled samples.
  \item \texttt{Uniform:} negative labels are selected uniformly over all classes.
  \item \texttt{NN:} We use the Nearest Neighbor (NN) method to the exclude the class of the NN, exclude four classes with the NNs, or to label with the class with the furthest NN.
  \item \texttt{Threshold:} refers to the method of section \ref{implementation1}
  \item \texttt{Oracle:} negative labels are selected uniformly over all wrong classes.
\end{itemize}

Selecting negative labels uniformly over all classes appears to hurt performance, suggesting that negative labels must be selected more carefully in the classification setting. 
\texttt{NN} methods appear to improve over purely supervised training, however the effectiveness is limited by long preprocessing times and the high dimensionality of the data. 

The method described in section \ref{implementation1}, listed here as \texttt{Threshold}, achieves superior test error in comparison to \texttt{NN} and \texttt{Uniform} methods. 
In particular, it is competitive with \texttt{Oracle - 1}, an oracle which labels each unlabeled sample with one negative label which the sample is not a class of. 

It is no surprise that \texttt{Oracle - 3} improves substantially over \texttt{Oracle - 1}, and it is not inconceivable to develop methods which can accurately select a small number of negative labels, and these may lead to even better results when combined with other SSL methods.

We stress that this is not a definitive list of methods to implement negative sampling in SSL, and our fast proposed method, when combined with other SSL, already improves over the state-of-the-art.

\begin{table}[!h]
\centering
\caption{Test error achieved by various $\text{NS}^3\text{L}$ techniques on F-MNIST with all but 2,000 labels removed. We use a small CNN trained for 50 epochs. Where applicable, the number after the dash indicates the number of negative labels per sample selected. }
\begin{tabular}{c|cc} \toprule
    F-MNIST & 2,000 \\ \midrule
    Supervised & 17.25 $\pm$ .22 \\
    Uniform - 1 & 18.64 $\pm$ .38 \\
    Uniform - 3 & 19.35 $\pm$ .33 \\
    Exclude class of NN - 1 & 17.12 $\pm$ .15 \\
    Exclude 4 nearest classes with NN - 1 & 17.13 $\pm$ .21 \\
    Furthest class with NN - 1 & 16.76 $\pm$ .15 \\
    Threshold $T = 0.03$ & 16.47 $\pm$ .18 \\
    Threshold $T = 0.05$ & 16.59 $\pm$ .19 \\
    Oracle - 1 & 16.37 $\pm$ .12 \\
    Oracle - 3 & 15.20 $\pm$ .66 \\
 \bottomrule
\end{tabular}
\label{tableNS3LResults}
\end{table}

\end{appendix}

\end{document}